\documentclass[conference]{IEEEtran}
\IEEEoverridecommandlockouts
\usepackage{booktabs}
\usepackage{cite}
\usepackage{amsmath,amssymb,amsfonts}
\usepackage{algorithmic}
\usepackage{graphicx}
\usepackage{textcomp}
\usepackage{xcolor}
\def\BibTeX{{\rm B\kern-.05em{\sc i\kern-.025em b}\kern-.08em
    T\kern-.1667em\lower.7ex\hbox{E}\kern-.125emX}}
\begin{document}

\title{Computation-Aware Event-to-Frame Reconstruction via Selective Attention \\
}

\author{\IEEEauthorblockN{Jingqian Wu}
\IEEEauthorblockA{
\textit{The University of Hong Kong}\\
Hong Kong, China \\
jingqianwu@connect.hku.hk}
\and
\IEEEauthorblockN{Yunbo Jia}
\IEEEauthorblockA{
\textit{Beijing University of} \\ \textit{Posts and Telecommunications}\\
Beijing, China \\
jybconcorde@163.com}
\and
\IEEEauthorblockN{Edmund Y. Lam}
\IEEEauthorblockA{
\textit{The University of Hong Kong}\\
Hong Kong, China \\
elam@eee.hku.hk}
}

\maketitle

\begin{abstract}
Event-to-frame (E2F) reconstruction bridges asynchronous event streams with frame-based vision pipelines, but existing methods often face a trade-off between reconstruction quality and computational efficiency. In this work, we propose an efficient E2F framework that emphasizes causal temporal modeling and computation-aware design. The architecture adopts a recurrent encoder–decoder to incrementally aggregate event information with compact hidden states. To improve robustness under fast motion and illumination variations, a selective context fusion strategy is introduced to integrate event-driven features with prior intensity cues. Within this fusion process, a lightweight hybrid attention mechanism enhances feature selectivity without relying on heavy attention operations. Experimental results on standard benchmarks demonstrate that the proposed approach achieves competitive reconstruction performance while maintaining a favorable balance between accuracy and model complexity.
\end{abstract}


\section{Introduction}
\label{sec: intro}
Event cameras represent visual information as asynchronous streams of brightness changes, offering distinct advantages such as high temporal resolution and wide dynamic range~\cite{rebecq2019high, wu2025sweepevgs, rebecq2019events, wu2024ev}. These properties make them attractive for scenarios involving fast motion or challenging lighting~\cite{wu2026dark}. However, the event-based sensing paradigm is inherently incompatible with conventional frame-based vision systems~\cite{9138762}, which motivates the need for event-to-frame (E2F) reconstruction as an intermediate representation.~\cite{rebecq2019high} For many practical applications, E2F models must achieve not only high reconstruction quality but also strong computational efficiency, as excessive model complexity limits their applicability in resource-constrained settings~\cite{7599576}.

\begin{figure}[!t]
	
	\centering
	
	\includegraphics[width=\linewidth,scale=1.0]{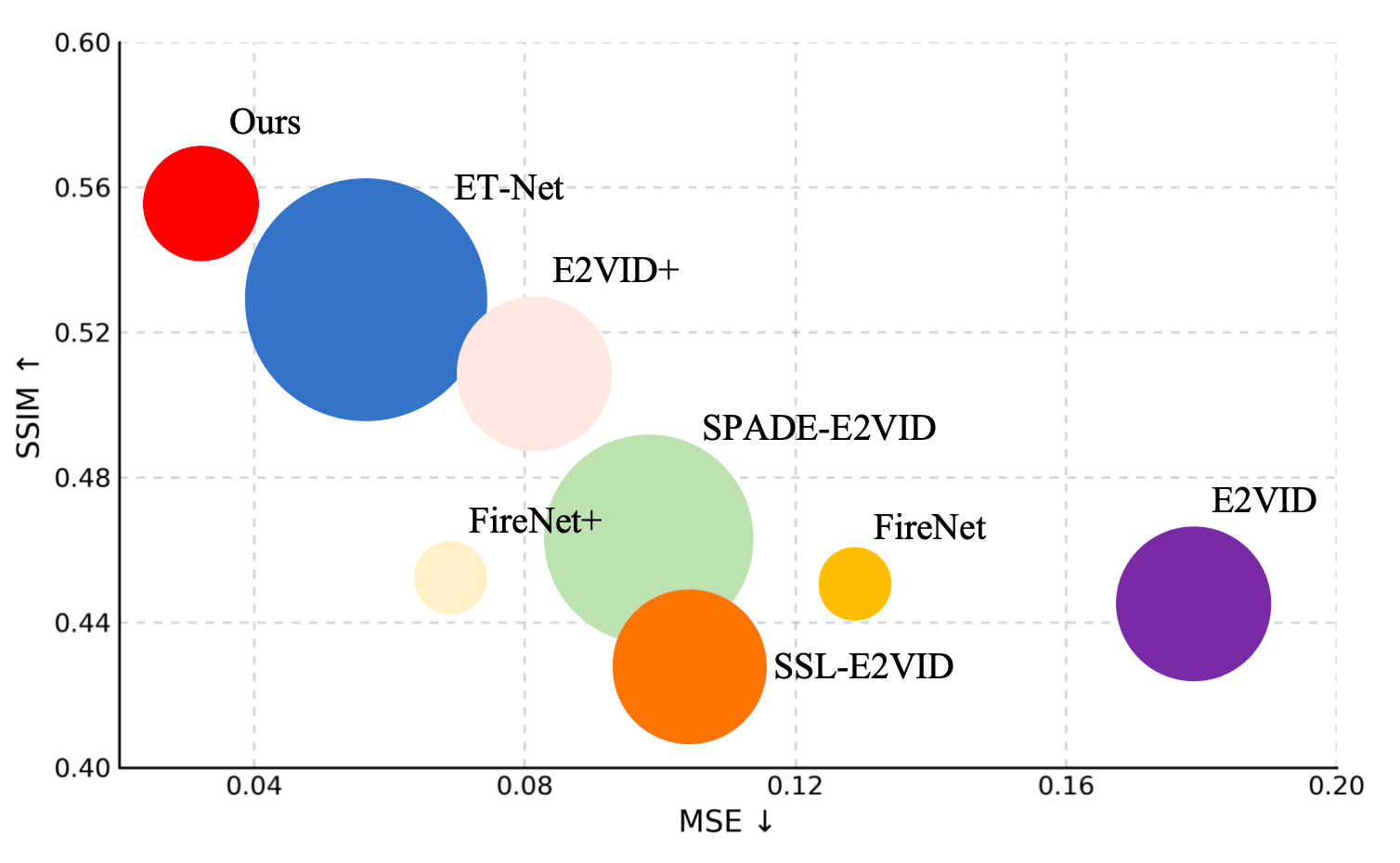}
	
	\caption{illustrates the quality-efficiency trade-off of event-to-frame (E2F) models on the Event Camera Dataset (ECD)~\cite{mueggler2017event}. Each scatter point represents reconstruction quality (SSIM ↑ vs. MSE ↓), while bubble size indicates the number of model parameters. Our method achieves the highest reconstruction quality with the smallest parameter footprint, demonstrating its suitability for scenarios requiring high fidelity under tight resource constraints.}
	
	\label{fig:teaser}
	
\end{figure}

Recent learning-based E2F approaches\cite{rebecq2019events,scheerlinck2020fast,stoffregen2020reducing,rebecq2019high,9337171} have significantly advanced reconstruction fidelity by leveraging recurrent architectures, multi-scale feature fusion, or attention-based mechanisms. While these designs improve temporal consistency and visual detail, they often introduce substantial computational overhead. In particular, transformer-style attention \cite{weng2021event} has been shown to enhance long-range dependency modeling, but at the cost of increased parameter count and multiply–accumulate operations (MACs). As a result, existing methods exhibit a clear quality–efficiency trade-off: lightweight models such as FireNet\cite{scheerlinck2020fast} and FireNet+ \cite{stoffregen2020reducing} offer favorable computational characteristics but limited representational capacity, whereas high-capacity models achieve strong reconstruction accuracy with considerably higher computational demands.

In this work, we focus on addressing this trade-off through a computation-aware E2F formulation that emphasizes compact temporal modeling and selective context integration. Our approach is built around a recurrent encoder–decoder architecture that performs causal temporal aggregation using convolutional recurrent units, providing effective temporal modeling with controlled complexity. To further enhance robustness under rapid motion and illumination variation, we incorporate a selective context fusion mechanism that integrates event-driven information with prior intensity content. Within this process, a lightweight hybrid attention strategy enables adaptive feature reweighting across spatial and temporal dimensions, improving reconstruction quality while remaining consistent with the lightweight design philosophy. Experimental results demonstrate that the proposed method achieves a favorable balance between reconstruction performance and computational efficiency when compared with existing baseline approaches.

\section{Related Work}
\subsection{Learning-based E2F Reconstruction.}
E2VID~\cite{rebecq2019events} pioneered learning-based event-to-frame (E2F) reconstruction by converting event streams into voxel grids and training a recurrent U-Net on large-scale simulated data, achieving improved reconstruction quality with good generalization to real events. A subsequent variant~\cite{rebecq2019high} replaced the recurrent unit with ConvLSTM to enhance temporal consistency across event window sizes. FireNet~\cite{scheerlinck2020fast} introduced a compact recurrent architecture that significantly reduces computational cost and memory footprint, albeit at the expense of reconstruction fidelity under complex motion and background conditions. Stoffregen \emph{et~al.}~\cite{stoffregen2020reducing} improved the stability of E2VID- and FireNet-based models by employing synthetic training data better matched to real event statistics. Cadena \emph{et~al.}~\cite{9337171} incorporated SPADE layers into E2VID to enhance contrast and detail. Departing from end-to-end RNN regression, Zhang \emph{et~al.}~\cite{ELRP} reformulated E2F reconstruction as a linear inverse problem coupling optical flow and brightness estimation, solved with classical and deep CNN-based regularizers for improved explainability. ET-Net~\cite{weng2021event} integrated Transformer blocks~\cite{vaswani2017attention} to model long-range dependencies, improving reconstruction fidelity while substantially increasing parameter count, memory usage, and inference cost, which limits applicability to embedded and neuromorphic platforms. More recently, diffusion-based approaches have emerged as a promising direction, from Event-Diffusion~\cite{eventdiffusion2023}, which first introduced denoising diffusion models to E2F reconstruction for improved edge restoration, to E2VIDiff~\cite{e2vidiff2024} and UniE2F~\cite{xu202Xunie2f}, which leverage pretrained generative priors and unified video diffusion frameworks to overcome the perceptual limitations of deterministic regression models.

\subsection{Lightweight and Realtime Neuromorphic System}

In circuit- and system-level deployments, strict latency constraints, limited on-chip SRAM, and the high energy cost of DRAM access make heavy attention-based backbones impractical, even when reconstruction quality is strong \cite{davies2018loihi, merolla2014million}. As a result, E2F methods face a quality–efficiency trade-off: lightweight models \cite{scheerlinck2020fast} are more deployable but often sacrifice accuracy, whereas high-fidelity approaches such as ET-Net \cite{weng2021event} incur prohibitive computational and memory overheads. In a similar efficiency-oriented vein, E2VIDX~\cite{E2VIDX} redesigned the E2VID backbone with group and sub-pixel convolutions, reducing model size by roughly 25\% while marginally improving SSIM/LPIPS, illustrating the value of architecture-level compression for bridging conventional and neuromorphic vision pipelines.

Motivated by practical constraints on latency, memory, and power, we adopt a hardware-aware and parameter-efficient design that combines \emph{causal} LSTM-based temporal aggregation with \emph{lightweight}, scene-adaptive dynamic filtering. By retaining only essential components—recurrent encoders for low-latency temporal modeling and compact conditioning mechanisms for context-dependent modulation—the proposed approach maintains strong reconstruction performance while aligning with neuromorphic deployment requirements.

\section{Methodology}
\subsection{Problem Formulation and Overview.}
\begin{figure*}[t]
	
	\centering
	
	\includegraphics[width=\linewidth,scale=1.0]{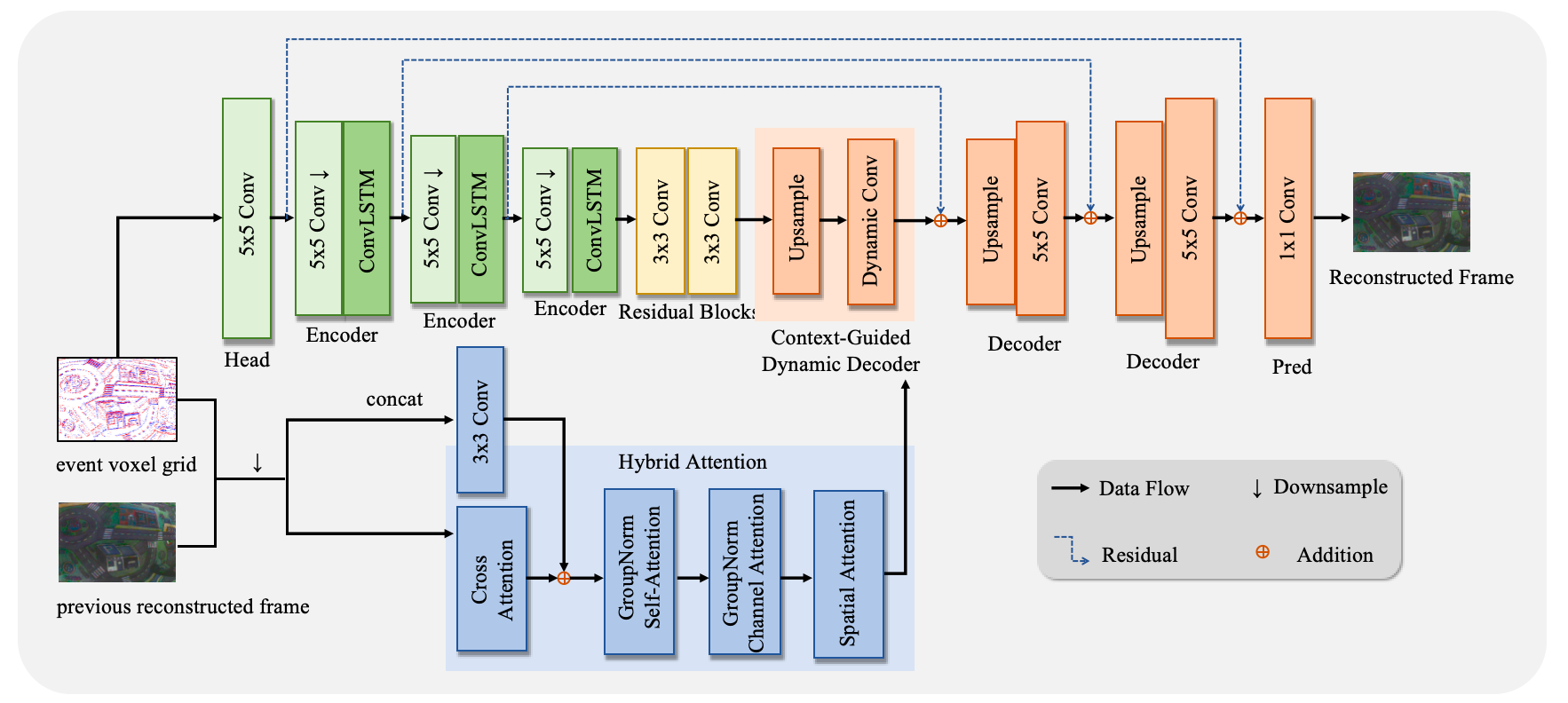}
	
	\caption{Overview of the proposed event-to-frame reconstruction framework. An input event voxel grid is first processed by a lightweight convolutional head and a recurrent encoder composed of strided convolutions and ConvLSTM units to perform causal temporal aggregation. A shallow residual block refines the encoded representation at the bottleneck. Context fusion is achieved by combining event-driven features with the previously reconstructed frame, which are jointly processed by a hybrid attention module to enable selective integration of dynamic and structural cues. The resulting context-aware representation conditions a dynamic decoding stage, followed by a convolutional decoder with additive skip connections that progressively restores spatial resolution.}
	
	\label{fig:pipeline}
	
\end{figure*}

Given a stream of events represented as voxelized tensors $\mathbf{V}_k \in \mathbb{R}^{H \times W \times B}$ at time step $k$, the goal of event-to-frame (E2F) reconstruction is to estimate the corresponding intensity image $\hat{\mathbf{I}}_k$. The proposed framework follows an encoder–decoder architecture with recurrent states, enabling causal temporal aggregation across consecutive event windows.  The overall architecture is illustrated in Figure~\ref{fig:pipeline}.

\subsection{Efficient Neuromorphic Vision Models.}
Beyond reconstruction accuracy, practical E2F deployment emphasizes computational efficiency and system-level constraints. Event-based vision systems often target platforms with limited memory bandwidth and power budgets, where model size and multiply–accumulate operations (MACs) directly affect feasibility. As a result, lightweight and computation-aware designs are central to neuromorphic and embedded vision research.

In this context, dynamic filtering and context-aware modulation have emerged as effective strategies for balancing performance and efficiency. These approaches suggest that selectively adapting features within convolutional pipelines can improve reconstruction quality while preserving a lightweight profile. Following this direction, our approach combines compact recurrent modeling with selective context fusion to meet the efficiency requirements of event-based vision systems.

\subsection{Recurrent Encoder–Decoder for Causal Modeling.}
Temporal consistency is achieved using a convolutional recurrent encoder–decoder that incrementally integrates event information over time. The encoder performs spatial downsampling followed by convolutional recurrent units to aggregate motion cues while maintaining compact hidden states, enabling causal temporal modeling without access to future events or long temporal buffers.

At the bottleneck, a shallow residual block refines the aggregated representation. The decoder then progressively restores spatial resolution through upsampling and convolutional refinement, with additive skip connections preserving fine-grained spatial details.

\subsection{Convolutional Context Fusion}
We introduce a convolutional context fusion mechanism that integrates features from the current event tensor and the previously reconstructed frame. Rather than uniformly fusing contextual cues, the proposed design selectively emphasizes signals most relevant to the current reconstruction step.

Let $\mathbf{E}_k$ denote event-driven features and $\mathbf{R}_{k-1}$ intensity-based features from the previous output. After spatial alignment, these features are concatenated to form a compact context representation that conditions the decoder, providing a unified interface for integrating transient motion cues and stable appearance priors.

\subsection{Hybrid Attention for Selective Context Integration}
To enhance the selectivity of context fusion, we introduce a lightweight hybrid attention mechanism that adaptively reweights fused features across modalities, spatial locations, and channels. 

\noindent\textbf{Cross-modal attention.}
Interactions between event-driven and intensity-based features are modeled via bidirectional cross-attention using pointwise convolutional projections and sigmoid gating:
\[
\mathbf{O}_{r \rightarrow e} = \mathbf{V}_e(\mathbf{E}) \odot 
\sigma(\mathbf{Q}_r(\mathbf{R}) \odot \mathbf{K}_e(\mathbf{E})),
\]
with a symmetric formulation for $\mathbf{O}_{e \rightarrow r}$. The two outputs are combined using normalized learnable weights to form a jointly conditioned representation.

\noindent\textbf{Efficient self-attention.}
To capture broader spatial context with linear complexity, a self-attention branch aggregates features through normalized key–value interactions:
\[
\mathbf{O}_{\text{sa}} = \mathbf{Q}(\text{softmax}(\mathbf{K})^\top \mathbf{V}),
\]
which is fused with a depthwise convolution branch to preserve local structure.

\noindent\textbf{Channel and spatial reweighting.}
Finally, channel-wise and spatial attention refine the fused representation using global pooling and lightweight convolutional gating, emphasizing informative feature dimensions with minimal overhead.

Overall, the hybrid attention module enables selective context integration within a convolutional fusion pipeline, improving reconstruction robustness while maintaining computational efficiency.

\begin{table*}[h]

\centering
\renewcommand{\arraystretch}{1.6}

\caption{Evaluation results of the existing methods and our proposed method on ECD and MVSEC datasets. Best results are in \textbf{bold}, second-best are \underline{underlined}.}

\begin{tabular}{lcccccccc}
\toprule
& & & \multicolumn{3}{c}{ECD\cite{mueggler2017event}}    & \multicolumn{3}{c}{MVSEC\cite{zhu2018multivehicle}} \\
\cmidrule(lr){4-6} \cmidrule(lr){7-9}

Method
& \#Params (M)
& MACs (G)
& \multicolumn{1}{c}{MSE$\downarrow$}
& \multicolumn{1}{c}{SSIM$\uparrow$}
& \multicolumn{1}{c}{LPIPS$\downarrow$}
& MSE$\downarrow$
& SSIM$\uparrow$
& LPIPS$\downarrow$ \\

\midrule  

E2VID\cite{rebecq2019events} & 10.71 & 42.24 & 0.179 & 0.450 & 0.322 & 0.225 & 0.241 & 0.645 \\
FireNet\cite{scheerlinck2020fast} & 0.038 & 3.63 & 0.133 & 0.459 & 0.321 & 0.294 & 0.198 & 0.702 \\
E2VID+\cite{stoffregen2020reducing} & 10.71 & 42.23 & 0.070 & 0.503 & 0.236 & 0.132 & 0.262 & \underline{0.514} \\
FireNet+\cite{stoffregen2020reducing} & 0.038 & 3.41 & 0.062 & 0.452 & 0.289 & 0.219 & 0.212 & 0.570 \\
SPADE-E2VID\cite{9337171} & 11.46 & 146.58 & 0.091 & 0.461 & 0.337 & 0.138 & \underline{0.266} & 0.591 \\
E2VIDX\cite{E2VIDX} & - & - & 0.060 & 0.450 & 0.324 & - & - & - \\
SSL-E2VID\cite{paredes2021back} & 10.71 & 42.23 & 0.092 & 0.415 & 0.380 & \underline{0.124} & 0.264 & 0.694 \\
ELRP\cite{ELRP} & - & - & 0.080 & 0.437 & 0.485 & - & - & - \\
ET-Net\cite{weng2021event} & 22.18 & 104.91 &{\underline{0.047}} & {\underline{0.552}} & {\underline{0.224}} & \textbf{0.107} & \textbf{0.288} & \textbf{0.489} \\
\textbf{Ours} & 10.16 & 37.83 & \textbf{0.034} & \textbf{0.554} & \textbf{0.224} & 0.127 & \underline{0.266} & 0.530\\
\bottomrule
\end{tabular}

\label{table:eval}
\end{table*}

\begin{figure*}[h]
  \centering
  \includegraphics[width=\linewidth]{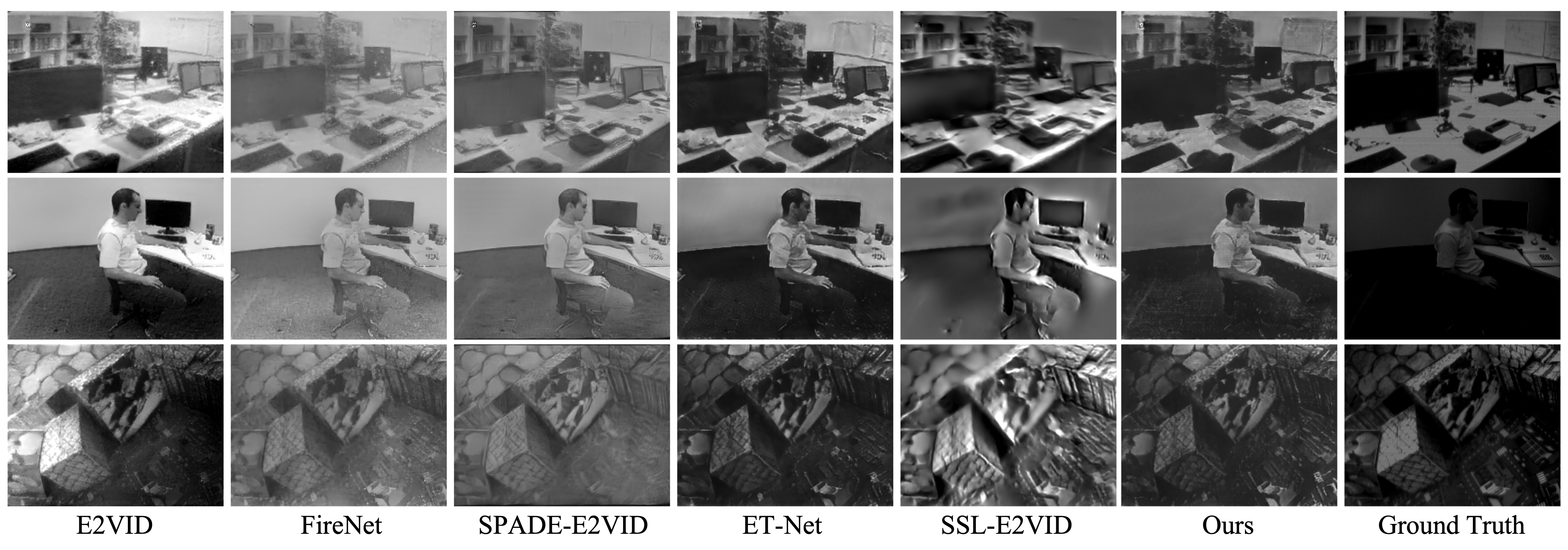} % 宽度匹配文本行宽
  \caption{Qualitative comparison of reconstructed intensity frames on real-world event sequences.
The proposed method demonstrates improved edge clarity, texture preservation, and brightness consistency in both indoor and outdoor environments. The resulting reconstructions exhibit closer agreement with the ground-truth images, suggesting enhanced structural fidelity under varying illumination conditions.
  } % 可选标题
  \label{fig:visualize}
\end{figure*}

\section{Experiments}
\subsection{Training Protocols}
Following prior work \cite{stoffregen2020reducing}, training data are synthesized by sampling natural images from MSCOCO \cite{lin2014microsoft} and rendering multi-object motion using the 2D renderer in ESIM \cite{rebecq2018esim}. MSCOCO images serve as both background and moving foreground content. The resulting dataset contains $280$ sequences of $10$\,s each, with foreground objects exhibiting diverse speeds and trajectories. For each sequence, event streams, reference intensity frames, and optical flow are generated at an average rate of $51$\,Hz. Both event and frame sensors use a resolution of $256{\times}256$, and the event contrast threshold is uniformly sampled from $[0.1,\,1.5]$.

\subsection{Evaluation Setup}
The proposed method is evaluated on standard benchmark datasets providing synchronized event streams and ground-truth intensity frames. All experiments use a voxel-grid representation for event data and fixed evaluation configurations to ensure fair comparison. Training and testing are conducted with identical input resolutions and preprocessing pipelines across all methods.

Reconstruction quality is evaluated using mean squared error (MSE), structural similarity index (SSIM), and learned perceptual image patch similarity (LPIPS), which together measure pixel-level accuracy, structural consistency, and perceptual similarity. Qualitative comparisons of 
reconstruction results across methods are presented in Figure~\ref{fig:visualize}.

\subsection{Quantitative Results}
Table~\ref{table:eval} reports quantitative results on the ECD and MVSEC datasets, with comparisons spanning lightweight recurrent models~\cite{scheerlinck2020fast,stoffregen2020reducing}, high-capacity convolutional baselines~\cite{rebecq2019events,stoffregen2020reducing,9337171}, self-supervised approaches~\cite{paredes2021back}, inverse-problem formulations~\cite{ELRP}, and transformer-based methods~\cite{weng2021event}.

\noindent\textbf{Performance on ECD.}
The proposed method achieves the best MSE of $0.034$ and SSIM of $0.554$, while matching ET-Net~\cite{weng2021event} on LPIPS ($0.224$). It reduces MSE by $51\%$ over E2VID+~\cite{stoffregen2020reducing} ($0.070\!\rightarrow\!0.034$) and by $27.7\%$ over ET-Net ($0.047\!\rightarrow\!0.034$), while using less than half the parameters and roughly one-third of the MACs of the latter, indicating that selective context fusion offers an effective alternative to global self-attention.

\noindent\textbf{Performance on MVSEC.}
Under fast ego-motion and complex outdoor illumination, our method attains an MSE of $0.127$, SSIM of $0.266$, and LPIPS of $0.530$. Although ET-Net~\cite{weng2021event} achieves the best absolute scores, our LPIPS outperforms all non-transformer baselines (e.g., $0.514$ for E2VID+~\cite{stoffregen2020reducing}, $0.694$ for SSL-E2VID~\cite{paredes2021back}), recovering most of the transformer's perceptual quality with substantially lower computational cost.

\noindent\textbf{Cross-dataset consistency and efficiency.}
Unlike baselines that exhibit dataset-specific behavior—e.g., FireNet+~\cite{stoffregen2020reducing} degrades sharply from ECD ($0.062$) to MVSEC ($0.219$), and SSL-E2VID~\cite{paredes2021back} suffers in perceptual quality (LPIPS $0.694$)—our method ranks first or second across all six metric–dataset combinations. Combined with its compact footprint (Figure~\ref{fig:teaser}), this places the proposed model on a favorable point of the quality–efficiency frontier, particularly relevant for resource-constrained deployment.

\subsection{Efficiency Analysis}
In addition to reconstruction quality, we evaluate model efficiency using parameter count and multiply–accumulate operations (MACs), which serve as hardware-agnostic indicators of computational complexity (Figure~\ref{fig:teaser}). The proposed model contains $10{,}156{,}254$ parameters and requires $37.8$\,G MACs per event voxel input, making it the most compact among methods achieving top-tier reconstruction quality.

\noindent\textbf{Comparison with transformer-based methods.}
Against ET-Net~\cite{weng2021event}, which represents the current state of the art, our model uses $54.2\%$ fewer parameters ($22.18\!\rightarrow\!10.16$\,M) and $63.9\%$ fewer MACs ($104.91\!\rightarrow\!37.83$\,G), yet achieves a lower MSE and comparable SSIM on ECD. This suggests that the proposed lightweight hybrid attention captures most of the benefits of global self-attention without its quadratic cost, which is critical for embedded and neuromorphic platforms with tight memory and energy budgets.

\noindent\textbf{Comparison with convolutional baselines.} Relative to backbones of similar capacity such as E2VID~\cite{rebecq2019events}, E2VID+~\cite{stoffregen2020reducing}, and SSL-E2VID~\cite{paredes2021back} ($\sim$$10.71$\,M params, $42.23$\,G MACs), our model is slightly smaller in both dimensions while delivering a $51\%$ MSE reduction over E2VID+ on ECD. SPADE-E2VID~\cite{9337171} further illustrates the cost of conditioning-heavy designs, requiring nearly $4\times$ our MACs without matching our accuracy. 

\noindent\textbf{Source of efficiency.}
This balance stems from two design choices: causal recurrent temporal modeling, which bounds per-step computation by avoiding long event-window buffering, and lightweight hybrid attention within a convolutional fusion pipeline, which improves feature selectivity via linear-complexity self-attention and pointwise cross-modal gating rather than dense global attention.

\begin{table}[h]
\centering
\renewcommand{\arraystretch}{1.6}
\caption{Ablation study of the fusion strategy in the Convolutional Context Fusion module on ECD and MVSEC datasets. Best results are in \textbf{bold}, second-best are \underline{underlined}.}
\resizebox{\columnwidth}{!}{%
\begin{tabular}{lcccccc}
\toprule
& \multicolumn{3}{c}{ECD\cite{mueggler2017event}} & \multicolumn{3}{c}{MVSEC\cite{zhu2018multivehicle}} \\
\cmidrule(lr){2-4} \cmidrule(lr){5-7}
Method
& \multicolumn{1}{c}{MSE$\downarrow$}
& \multicolumn{1}{c}{SSIM$\uparrow$}
& \multicolumn{1}{c}{LPIPS$\downarrow$}
& MSE$\downarrow$
& SSIM$\uparrow$
& LPIPS$\downarrow$ \\
\midrule
Addition           & 0.040 & 0.524 & 0.251 & \textbf{0.123} & 0.256 & 0.568 \\
Concatenation      & 0.040 & 0.524 & 0.250 & \underline{0.125} & 0.255 & 0.554 \\
Lightweight Gating & \underline{0.035} & \underline{0.538} & \underline{0.243} & 0.127 & \underline{0.265} & \underline{0.552} \\
\textbf{Hybrid Attention (Ours)}
                   & \textbf{0.034} & \textbf{0.554} & \textbf{0.224}
                   & 0.127           & \textbf{0.266} & \textbf{0.530} \\
\bottomrule
\end{tabular}%
}
\label{table:ablation_fusion}
\end{table}

\subsection{Ablation Study}
To validate the effectiveness of the proposed hybrid attention design in our Convolutional Context Fusion module, we replace it with three progressively simpler fusion alternatives while keeping all other components of the network unchanged: (i) \textbf{Addition}, in which the downsampled event and previous-reconstruction features are projected and element-wise added to the convolved context; (ii) \textbf{Concatenation}, in which the three streams are concatenated and fused by 1$\times$1 convolutions; and (iii) \textbf{Lightweight Gating}, an SE-style channel-wise gating mechanism that adaptively reweights the projected event and reconstruction branches before summation. These three variants form a complexity gradient from purely linear fusion to channel-selective fusion, allowing us to isolate the contribution of cross-modal, self, channel, and spatial attention in our full design. All variants are trained under identical settings and evaluated on the ECD and MVSEC datasets using MSE, SSIM, and LPIPS.

As shown in Table~\ref{table:ablation_fusion}, the proposed hybrid attention consistently outperforms the simpler baselines on the ECD dataset across all three metrics, reducing MSE from 0.040 to 0.034, improving SSIM from 0.524 to 0.554, and lowering LPIPS from 0.251 to 0.224. The Addition and Concatenation variants yield nearly identical results, indicating that without an explicit selection mechanism the network struggles to disentangle complementary information from the event stream and the recurrent image prior. The Lightweight Gating variant narrows this gap by introducing channel-wise selectivity, but it still falls short of the proposed design, suggesting that channel reweighting alone is insufficient and that joint modeling of cross-modal, spatial, and global dependencies is essential for high-quality reconstruction. On the MVSEC dataset, the proposed method achieves the best SSIM (0.266) and a competitive MSE (0.127), while the simpler variants attain marginally lower MSE values; however, this comes at the cost of noticeably degraded perceptual quality, as reflected by the higher LPIPS scores (0.568 and 0.554 for Addition and Concatenation, respectively, versus 0.530 for ours). This is consistent with the well-documented perception–distortion trade-off in event-based video reconstruction, and confirms that the hybrid attention module produces reconstructions that are both structurally faithful and perceptually closer to the ground truth.

\section{Conclusion}
This paper presented an efficient event-to-frame reconstruction framework that combines causal recurrent modeling with selective context fusion. By embedding a lightweight hybrid attention mechanism within a convolutional fusion pipeline, the proposed approach adaptively emphasizes informative features while preserving computational efficiency. Extensive evaluations on standard benchmarks demonstrate competitive reconstruction performance and a favorable balance between accuracy and model complexity. Overall, this work highlights the effectiveness of selective, convolution-friendly attention for end-to-end event-based reconstruction under constrained computational budgets.

\section*{Acknowledgments}
This work is supported by the Research Grants Council of Hong Kong (GRF 17201822) and the Theme-based Research Scheme (T45-701/22-R).

\bibliography{references}
\bibliographystyle{IEEEtran}

\end{document}